\documentclass[10pt,technote,twoside,twocolumn]{IEEEtran}

\usepackage[cmex10]{amsmath}
\usepackage{times}
\usepackage{graphicx,epstopdf}
\usepackage{amsmath,eucal,amssymb}
\usepackage{amssymb,bibspacing,url,xspace}
\usepackage[bf,small]{caption}

\usepackage{mdwtab}
\usepackage{multirow}

\usepackage[small]{subfigure}

\usepackage[]{cite}
\usepackage{comment}

\usepackage{algorithm2e}
\let\savedalgorithm\algorithm
\let\savedendalgorithm\endalgorithm

\usepackage{algorithm}

\def\onedot{.}
\def\eg{\emph{e.g}\onedot}
\def\ie{\emph{i.e}\onedot}

\def\etal{\emph{et al}\onedot}
\def\half{{\tfrac{1}{2} }}
\def\argmax{\operatorname*{argmax\,}}
\def\argmin{\operatorname*{argmin\,}}
\def\T{{\!\top}}

\def\Real{\mathbb{R}}

\def\dt{{\bf H}}
\def\dtree{{\hbar}}

\def\bw{{\bf w}}
\def\bx{{\bf x}}
\def\by{{\bf y}}

\def\bh{{ \Phi }}

\let\x\bx
\let\y\by
\let\w\bw
\let\c\bc

\def\z{{\bf y}'}

\def\blambda{{\boldsymbol \lambda}}
\def\bxi{{\boldsymbol \xi}}

\def\calY{{\cal Y}}

\def\calH{ {\cal H } }

\def\loss{{\it \Delta}}
\def\bh{{\Phi }}

\def\btheta{{\boldsymbol \theta}}

\newcommand{\st}{{{\rm s.t.}\!:}\xspace}
\newcommand{\fnorm}[2][2]{\ensuremath{ \left\| #2 \right\|_{ \mathrm{#1} } } }

\newcommand{\structens}{{CRFT}ree\xspace}
\newcommand{\ssvm}{{SSVM}s\xspace}

\newcommand{\structsvm}{{SSVM}s\xspace}
\newcommand{\svm}{{SVM}s\xspace}

\newcommand{\adaboost}{{AdaBoost}\xspace}
\newcommand{\crf}{{CRF}s\xspace}
\newcommand{\crfs}{{CRF}s\xspace}

\newcommand{\oneslack}{{$ 1 $-slack}\xspace}

\let\sens\structens

\def\paragraph{\textbf}

\begin{document}

\title{Structured Learning of Tree Potentials in CRF for Image Segmentation}

\author{Fayao~Liu, Guosheng~Lin, Ruizhi~Qiao, Chunhua~Shen
\IEEEcompsocitemizethanks{
\IEEEcompsocthanksitem
F. Liu, R. Qiao, C. Shen are with The University of Adelaide,
Australia.
G. Lin is with Nanyang Technological University,
Singapore. This work was done when G. Lin was with The University of Adelaide.
Email: \{fayao.liu, ruizhi.qiao, chunhua.shen\}@adelaide.edu.au,
guosheng.lin@gmail.com

\it Appearing in IEEE Transactions on Neural Networks and Learning Systems, 26 March 2017.
}

}

\maketitle

\begin{abstract}

    We propose a new approach to image segmentation, which exploits the advantages of both
    conditional random fields (\crf) and decision trees.
	In the literature, the potential functions of \crf are mostly defined as a \emph{linear} combination of some pre-defined parametric models, and then methods like structured support vector machines (\ssvm) are applied to learn those linear coefficients. We instead formulate the unary and pairwise potentials as nonparametric forests---ensembles of decision trees, and learn the ensemble parameters and the trees in a unified optimization problem within the large-margin framework. In this fashion, we easily achieve \emph{nonlinear} learning of potential functions on both unary and pairwise terms in \crf. Moreover, we learn class-wise decision trees for each object that appears in the image.
    Due to the rich structure and flexibility of decision trees, our approach is powerful in modelling complex data likelihoods and label relationships.
    The resulting optimization problem is very challenging because
    it can have exponentially many variables and constraints.
    We show that this challenging optimization can be efficiently solved by
    combining a modified column generation and cutting-planes
    techniques.
    Experimental results on both binary (Graz-02, Weizmann horse, Oxford flower) and multi-class (MSRC-21, PASCAL VOC 2012)
    segmentation datasets demonstrate the power of the learned nonlinear nonparametric potentials.

\end{abstract}


\tableofcontents
\clearpage

\section{Introduction}

The goal of object segmentation is to produce a pixel level segmentation of different
object categories. It is challenging as the objects may appear in various backgrounds and
in different visual conditions.
\crfs \cite{Lafferty01Conditional}
model the conditional distribution of labels given observations, and represents the state-of-the-art
in image/object segmentation  \cite{SzummerKH08,Shotton08,Fulkerson09,Lucchi12,NowozinGL10,Kim2016}.
The max-margin principle has also been applied to predict structured outputs, including
\ssvm \cite{Tsochantaridis05}, and
        max-margin Markov networks \cite{TasGueKol03}.
        These three methods share similarities when viewed as optimization problems using
different loss functions.
Szummer \etal\ \cite{SzummerKH08} proposed to learn linear coefficients of \crf potentials using \ssvm and graph cuts.
To date, most of these methods assume a pre-defined parametric model for the potential functions,
and typically only the linear coefficients of the parametric model are learned.
This can greatly limit the flexibility of the model capability of \crf, and thus
calls for effective methods to incorporate nonlinear nonparametric models for learning the
potential functions in \crf.

    As similar in standard support vector machines (\svm),  nonlinearity can be achieved by introducing nonlinear kernels for \ssvm.
    However, the time complexity of nonlinear \svm is roughly $ O(n^{3.5}) $ with $ n $ being the number of training data examples.
    This time complexity is problematic for \ssvm, where the number of constraints grows exponentially in
    the description length of the label $ \y $. Moreover, nonlinear functions can significantly slow down
    the test time in most cases.
    Because of these reasons, currently most \ssvm applications use linear kernels
    (or linear parametric potential functions in \crf),
    despite the fact that nonlinear
    functions usually deliver more promising prediction accuracy.
    In this work, we address this issue by combining \crf with nonparametric decision trees.
    Both \crf and decision trees have gained tremendous success in computer vision.
    Decision trees are capable of modelling complex relations and generalize well on test data.
    Unlike kernel methods, decision trees are fast to evaluate and can be used to select
    informative features.

    In this work,
    we propose to use ensembles of decision trees to map the image content to both
    the unary terms and the pairwise interaction values in  \crfs. The proposed method is
    termed as \structens.
	Specifically, we formulate both the unary and pairwise potentials as \emph{nonparametric} forests---ensembles of decision trees, and learn the ensemble parameters and the trees in a single optimization framework. In this way, the nonlinearity is easily introduced into \crf learning without confronting the kernel dilemma.
	Furthermore, we learn class-wise decision trees for each object.
    Due to the rich structure and flexibility of decision trees, our approach is
    powerful in modelling complex data likelihoods and label relationships.
    The resulting optimization problem is very challenging in the sense that
    it can involve exponentially or even infinitely many variables and constraints.
    We summarize our main contributions as follows.
    \begin{itemize}
    \itemsep -1pt
    \item[1.]
	We formulate the unary and pairwise potentials as ensembles of decision trees, and show how to jointly learn the ensemble parameters and the trees as a unified optimization problem within the large-margin framework. In this fashion, we achieve nonlinear potential learning on both the unary and pairwise terms.
	\item[2.]
	We learn class-wise decision trees (potentials) for each object that appears in the image.
    \item[3.]
        We show how to train the proposed \structens model efficiently.
        In particular, we combine the column generation and cutting-planes techniques to approximately
        solve the resulting optimization problem, which can involve exponentially many variables and
        constraints.
    \item[4.]
        We empirically demonstrate that \structens outperforms  existing methods
        for image segmentation.
        On both binary  and multi-class
        segmentation datasets we show the advantages
        of the learned nonlinear nonparametric potentials of decision trees.

    \end{itemize}

{\bf Related work}
We briefly review the recent works that are relevant to ours.
A few attempts have been made to apply nonlinear kernels in \ssvm. Yu \etal \cite{Yu08} and Severyn \etal \cite{Severyn11} developed sampled cuts based methods for
training \ssvm with kernels.
Sampled cuts methods were originally proposed for standard kernel \svm. When
applied to \ssvm, the performance is compromised \cite{Lucchi12}.
In \cite{Bertelli11}, the image-mask pair kernels are designed to exploit image-level
structural information for object segmentation. However, these kernels are restricted to the unary term.
Although not in the large margin framework, the kernel \crf proposed in \cite{kcrf} incorporates kernels into the \crf learning.
The authors only demonstrated the efficacy of their method on a synthetic and a small scale protein dataset.
To sum up, these approaches are hampered by the heavy computation complexity.
Furthermore, it is not a trivial task to design appropriate kernels for structured problems.
Recently, Lucchi et al.\ \cite{Lucchi12} proposed a two-step solution to tackle this problem.
Specifically, they train linear \structsvm by using kernelized feature vectors that are
obtained from training a standard non-linear kernel \svm model. They experimentally demonstrate
that the kernel transferred linear \svm model achieves similar performance as the Gaussian \svm.
However, this approach is heuristic and it cannot be shown theoretically
that their formulation approximates a nonlinear
\ssvm model. Besides, their method consumes extra usage of memory and training time
since the dimension of the transformed features equals to the number of support vectors, while the
latter is linearly proportional to the size of the training data \cite{Steinwart2003}. Moreover, compared to the above mentioned works of \cite{Bertelli11} and \cite{Lucchi12}, we achieve nonlinear learning on both the unary and the pairwise terms while theirs are limited to nonlinear unary potential learning.
The recent work of Shen \etal \cite{structboost} generalizes standard boosting methods to structured learning, which shares similarities to our work here.
However, our method bears critical differences from theirs:
1) We design a column generation method for non-linear tree potentials learning in \crf directly from
the \ssvm formulation.
Different from the case in \cite{structboost}, which can directly derive column generation method analogous to LPBoost \cite{lpboost},
our derivation here is more challenging.
This is because we can not obtain the most violated constraint from the constraints of the dual problem, on which the column generation
technique relies. We instead inspect the KKT condition to seek for the most violated constraint.
This is an important difference compared to existing column generation techniques.
2) We develop a \crf learning method for multi-class semantic segmentation, while \cite{structboost} only shows
\crf learning for binary foreground/background segmentation.
Our experiments on the MSRC-21 dataset shows that our method achieves state-of-the-art results.
3) We learn class-wise decision trees (potentials) for each object that appears in the image.
This is different from \cite{structboost}.
 	The work of decision tree fields \cite{DTF} is close to ours in that they also use decision trees to model the
pairwise potentials.
The major difference is that in \cite{DTF} potential functions are  constructed  by directly summing the energy tables associated with the set of nodes taken during evaluating the decision trees. Their trees are generally deep, with depth 15 for the unary potential and 6 for the pairwise potential in their experiment.
By contrast, we model the potential functions as an ensemble of decision trees and learn them in the large margin framework.
In our method, the decision trees are shallow and simple  with binary  outputs.

\section{Learning tree potentials in \crf}
\label{sec:main}
We present the details of our method in this section by first introducing the \crf models for segmentation, then formulating the energy functions and showing how to learn decision tree potentials in the large-margin framework.

\subsection{Segmentation using \crf models}

Before presenting our method, we first revisit how to use \crf models to perform image segmentation.
Given an image instance $\x$ and its corresponding labelling $\y$,
\crf \cite{Lafferty01Conditional} models the conditional distribution of the form
\begin{align}  \label{eq:crf_log}
P(\y|\x; \w) = \frac{1}{Z} \exp (- E(\y, \x;\w)).
\end{align}
where  $\w$ are parameters and $Z$ is the normalization term.
The energy $E$ of an image $\x$ with segmentation labels $\y$ over the nodes (superpixels) $\cal N$ and edges $\cal S$, takes the following form:
\begin{align} \label{eq:seg_energy}
	E(\y, \x; \w)& =\sum_{p \in {\cal N} } \bh^{(1)}(y^{p}, \x;\w)
	 + \sum_{(p,q) \in {\cal S}} \bh^{(2)}(y^{p}, y^{q}, \x; \w).
\end{align}
Here $ \bx \in {\cal X}, \by \in {\cal Y}$;
$\bh^{(1)}$ and $\bh^{(2)}$ are the unary and pairwise potentials, both of which depend on the observations as well as the parameter $ \w $.
\crf seeks an optimal labeling that achieves maximum a posterior (MAP), which mainly involves a two-step process \cite{SzummerKH08}: 1) Learning the model parameters from the training data; 2) Inferring a most likely label for the test data given the learned parameters.
The segmentation problem thus reduces to minimizing the energy (or cost) over $\y$ by the learned parameters $ \w $, which is $\y^{*}=\argmin_{ \y \in  {\cal Y}} E(\y, \x; \w)$. When the energy function is submodular, this inference problem can be efficiently solved via graph cuts \cite{SzummerKH08}.

\subsection{Energy Formulation} \label{sec: energy}
Given the energy function in Eqn. \eqref{eq:seg_energy}, we show how to construct the unary and pairwise potentials using decision trees.
We denote $\x^{p}$ as the features of superpixel $p$ ($p=1, \ldots, n$), with its label $y^{p} \in \{ 1, \ldots, K\}$, where $K$ is the number of classes.
Let $\cal H$ be a set of decision trees, which can be infinite. Each $\dtree_j^{(1)}(\cdot) \in \cal H$ takes $\x^{p}$ as the input, and $\dtree_j^{(2)}(\cdot, \cdot) \in \cal H$ takes a pair $(\x^{p}, \x^{q})$ as the input to output $\{0, 1\}$.
We introduce $(K+1)$ groups of decision trees, in which $K$ groups are for the unary potential and one group for the pairwise potential.
For the unary potential, the $K$ groups of decision trees are denoted by $\dt_{c}^{(1)} (c=1,\ldots, K)$, which correspond to $K$ categories.
Each $\dt_{c}^{(1)}$ is associated with the $c$-th class.
In other words, for each class, we maintain its own unary feature mappings.
Each group of decision trees for the unary potential can be written as:
$\dt_{c}^{(1)}=[\dtree_{c1}^{(1)}, \dtree_{c2}^{(1)}, \ldots]^{\T}$, which are the output of decision trees: $\dtree_{cj}^{(1)}$.
All decision trees of the unary potential are denoted by
$\dt^{(1)}=[\dt_{1}^{(1)}, \dt_{2}^{(1)}, \ldots, \dt_{K}^{(1)}]$.
Accordingly, for the pairwise potential, the group of decision trees is denoted by $\dt^{(2)}$,
and $\dt^{(2)}=[\dtree_1^{(2)}, \dtree_{2}^{(2)}, \ldots]^{\T}$ being the output of all $\dtree_{j}^{(2)}$.
The whole set of decision trees is denoted by $\dt = [\dt^{(1)}, \dt^{(2)}]$.
We then construct the unary and pairwise potentials as
\begin{equation} \label{eq:unary}
       \bh^{(1)}(y^{(p)}, \x )  = \w_{y^p}^{(1)\T} \dt_{y^p}^{(1)}(\x^{p}).
\end{equation}
\begin{equation} \label{eq:pairwise}
       \bh^{(2)}( y^{(p)}, y^{(q)}, \x)  = \w^{(2)\T} \dt^{(2)}(\x^{p}, \x^{q}) I (y^{p} \neq y^{q} ).
\end{equation}
where $I(\cdot)$ is an indicator function which equals $1$ if the input is true and $0$ otherwise. Then the energy function in Eqn. \eqref{eq:seg_energy} can be written as:
\begin{align} \label{eq:seg_energy_dt}
	E(\y, \x; \w, \dt)& = \sum_{p \in {\cal N} } \w_{y^p}^{(1)\T} \dt_{y^p}^{(1)}(\x^{p})  \notag \\
	&+  \sum_{(p,q) \in {\cal S}} \w^{(2)\T} \dt^{(2)}(\x^{p}, \x^{q}) I (y^{p} \neq y^{q} ).
\end{align}
Next we show how to learn these decision tree potentials in the large-margin framework.

\subsection{Learning \crf in the large-margin framework}
Instead of directly minimizing the negative log-likelihood loss, we here learn the \crf parameters in the large margin framework, similar to \cite{SzummerKH08}.
Given a set of training examples $\{\x_i,\y_i\}_{i=1}^m$,
 the large-margin based \crf learning solves the following optimization:
\begin{align}
  \min_{\bw, \bxi \geq 0}   \;\; &
        \half \fnorm{\bw}^2 +  {\tfrac{C}{m}} \, \sum_i \xi_i \notag \\
  \st  \;\; &
    E(\y, \x_i; \w, \dt)- E (\y_i, \x_i; \w, \dt )\geq
    \loss ( \y_i, \y) - \xi_i, \notag \\
   & \quad\quad\quad
   \forall i=1, \dots, m,
   \text{ and }\forall \y \in  {\cal Y};.
\label{eq:ssvm_crf}
\end{align}
where $\loss: \calY \times \calY \mapsto \Real$ is a loss function associated with the prediction and the true label mask.  In general, we have $ \loss( \y, \y ) = 0$ and $ \loss(\y, \z ) > 0 $ for any $ \z \neq \y $.
Intuitively, the optimization in Eqn. \eqref{eq:ssvm_crf} is to encourage the
energy of the ground truth label $E(\y_i,\x_i; \w)$ to be lower than that of any other
{\em incorrect} labels $E(\y, \x_i; \w)$ by at least a margin $\loss (
\y_i, \y)$.

To learn the potential functions we proposed in \S\ref{sec: energy} in the large-margin framework, we introduce the following definitions.
For the unary part, we define $\w^{(1)}  = \w_1 ^{(1)} \odot \w_2 ^{(1)} \odot \ldots \odot \w_K^{(1)}$, where $\odot$ stacks two vectors, and
\begin{align} \label{eq:Psi1}
       \Psi^{(1)}(\y, \x; \dt^{(1)})  = \sum_{p \in {\cal N} } \dt_{y^p}^{(1)}(\x^{p}) \otimes y^p.
\end{align}
where $\otimes$ denotes the tensor operation (\eg, $\x^{p} \otimes y^p = [I(y^p=1)\x^{p\T}, \ldots, I(y^p=K)\x^{p\T}] ^{\T}$). Recall that $\x^{p}$ denotes the $p$-th superpixel of the image $\x$.
Here, $\Psi^{(1)}$ acts as the unary feature mapping.
Clearly we have:
\begin{align}  \label{eq:Psi1-Phi1}
       \w^{(1)\T} \Psi^{(1)}(\y, \x; \dt^{(1)}) = \sum_{p \in {\cal N} } \bh^{(1)}(y^p, \x).
\end{align}
For the pairwise part, we define the pairwise feature mapping as:
\begin{align} \label{eq:Psi2}
       \Psi^{(2)}(\y, \x; \dt^{(2)})  = \sum_{(p,q) \in {\cal S}} \dt^{(2)}(\x^{p}, \x^{q}) I (y^{p} \neq y^{q} ).
\end{align}
Then we have the following relation:
\begin{align} \label{eq:Psi2Phi2}
       \w^{(2)\T} \Psi^{(2)}(\y, \x; \dt^{(2)}) = \sum_{(p,q) \in {\cal S}} \bh^{(2)}(y^p, y^q, \x).
\end{align}
We further define $ \w  = \w ^{(1)} \odot \w^{(2)}$, and the joint feature mapping as
\begin{align} \label{eq:Psi}
       \Psi (\y, \x; \dt) = \Psi^{(1)}(\y, \x; \dt^{(1)}) \odot \Psi^{(2)}(\y, \x; \dt^{(2)}).
\end{align}
With the definitions of $\w$ and $\Psi$, the energy function can then be written as:
\begin{align} \label{eq:seg_energy_wPhi}
	E(\y, \x; \w, \dt)& =\sum_{p \in {\cal N} } \bh^{(1)}(y^{p}, \x;\w, \dt^{(1)})  \notag \\
	& \quad + \sum_{(p,q) \in {\cal S}} \bh^{(2)}(y^{p}, y^{q}, \x; \w, \dt^{(2)})  \notag \\
	& = \w^{\T} \Psi(\y, \x; \dt).
\end{align}

Now we can apply the large-margin framework to learn \crf using the proposed energy functions by rewriting the optimization problem in Eqn. \eqref{eq:ssvm_crf} as:
\begin{align}  \label{eq:ssvm_dt}
  \min_{\bw, \bxi}   \;\; &
        \half \fnorm{\bw}^2 +  {\tfrac{C}{m}} \, \sum_i \xi_i \notag \\
  \st  \;\; &
    \w^{\T} \left[ \Psi(\y, \x_i ; \dt) - \Psi(\y_i, \x_i ; \dt) \right] \geq
    \loss ( \y_i, \y) - \xi_i, \notag \\
   & \quad\quad\quad
   \forall i=1, \dots, m,
   \text{ and }\forall \y \in  {\cal Y}; \notag \\
   &\w \geq 0, \bxi \geq 0.
\end{align}
Note that we add the $\w \geq 0$ constraint to ensure submodular property of our energy functions, which we will discuss the details later in \S\ref{sec:one_slack}.
Up until now, we are ready to learn $\w$ and $\Psi$ (or $\dt$) in a single optimization problem formulated in Eqn. \eqref{eq:ssvm_dt}, but it is not clear how.
Next we demonstrate how to solve the optimization problem in Eqn. \eqref{eq:ssvm_dt} by using column generation and cutting-plane.

\subsection{Learning tree potentials using column generation}
\label{sec:main_algo}
We aim to learn a set of decision trees $\dt$ and the potential parameter $\bw$ by solving the
optimization problem in Eqn. \eqref{eq:ssvm_dt}.
However, jointly learning $\dt$ and $\bw$ is generally difficult.
Here we propose to apply column generation techniques \cite{lpboost,multiboost} to alternatively construct the set of decision trees and solve for $\bw$.
From the point of view of column generation techniques,
the dimension of the primal variable $\bw$ is infinitely large;
the column generation is to iteratively select (generate) variables for solving the optimization.
In our case, infinitely many dimension of $\bw$ corresponds to infinitely many decision trees,
thus we iteratively generate decision trees to solve the optimization.

Basically, we construct a working set of decision trees (denoted as $\mathcal W_\dt$).
During each column generation iteration we perform two steps.
In the first step, we generate new decision trees and add them to $\mathcal W_\dt$.
In the second step,
we solve a restricted optimization problem in Eqn. \eqref{eq:ssvm_dt} on the current working set $\mathcal W_\dt$ to
obtain the solution of $\bw$.
We repeat these two steps until convergence.
Next we describe how to generate decision trees in a principal way by using the dual solution of the optimization in Eqn. \eqref{eq:ssvm_dt}, which is similar to the conventional column generation technique.
First we derive the Lagrange dual problem of Eqn. \eqref{eq:ssvm_dt}, which can be written as
            \begin{eqnarray}  \label{eq:dual}
        \begin{split}
            \max_{\blambda, \btheta}   \;\; &
            \sum_{i, \y } \lambda_{ (  i, \y) } \loss( \y_i, \y )  \\
            &- \half \biggl\{ \sum_{i, \y } \lambda_{ (  i, \y) }
            \left[ \Psi(\y, \x_i; \dt) - \Psi(\y_i, \x_i; \dt) \right]
            + {\boldsymbol \theta}\biggr\}^2
             \\
            \st \;\; &  0 \leq \textstyle \sum_{ \y } \lambda_{ (  i, \y)
                          } \leq
                          \tfrac{C}{m}, \forall i=1,\dots, m; \btheta \geq 0, \blambda \geq 0.
        \end{split}
\end{eqnarray}
        Here ${\boldsymbol \theta}, \boldsymbol  \lambda $ are the dual variables.
        When using column generation technique, one need to find the most violated constraint in the dual. However, the constraints of the dual problem do not involve decision trees $\dt$. Instead of examining  the dual constraint, we inspect the KKT condition, which is an important difference compared to existing column generation techniques.
        According to the KKT condition, when at optimal, the following condition holds for the primal solution $ \w $ and the current working set $\mathcal W_{\dt}$:
        \begin{align} \label{eq:KKTw}
             \w \geq \sum_{i, \y  } \lambda_{ (  i, \y) } [ \Psi(\y, \x ; \dt) - \Psi(\y_i, \x ; \dt)].
        \end{align}
        All of those generated $\dt \in \mathcal W_{\dt}$ satisfy the above condition.
        Obviously, generating new decision trees which most violate the above condition would contribute the most to the optimization of Eqn. \eqref{eq:ssvm_dt}. Hence the strategy of generating new decision trees is to solve the following problem:
	     \begin{equation} \label{eq:cg_all}
        		\dt^{\star} = \argmax_{\dt} \sum_{i, \y } \lambda_{ (  i, \y) }
        [ \Psi (\y, \x_i; \dt) - \Psi (\y_i, \x_i; \dt)].
        \end{equation}
        Then $\dt^{\star}$ is added to the current working set $\mathcal W_{\dt}$.
        If $\dt^{\star}$ still satisfies the condition in Eqn. \eqref{eq:KKTw},
        the current solution of $\dt$ and $\bw$ is already the globally optimal one.

        The optimization in Eqn. \eqref{eq:cg_all} for generating new decision trees can be independently decomposed into solving the unary part and the pairwise part.
        Hence $\dt^{\star}$ can be written as: $\dt^{\star} = [\dt^{(1)\star}, \dt^{(2)\star}]$.
         For the unary part, we learn class-wise decision trees, namely, we generate $K$ decision trees corresponding to $K$ categories  at each column generation iteration.
         Hence $\dt^{(1)\star}$ is composed of $K$ decision trees: $\dt^{(1)\star}=[\dtree_1^{(1)\star}, \dots, \dtree_K^{(1)\star}]$.
                More specifically, according to the definition of $\Psi(\y, \x)$ in Eqn. \eqref{eq:Psi}, we solve the following $K$ problems:
       \begin{align}  \label{eq:dtree1}
          \forall c=&1,\ldots, K: \notag \\
           \dtree_c^{(1)\star}(\cdot) & = \argmax_{ \dtree \in \calH } \;
                         \sum_{i, \y  }  \lambda_{ ( i, \y ) }
                            \biggr[ \sum_{p \in {\cal N}, \atop y^p=c } \dtree_{y^p}^{(1)}(\x_i^{p}) - \sum_{p \in {\cal N}, \atop y_i^p=c } \dtree_{y_i^p}^{(1)}(\x_i^{p}) \biggr] \notag \\
                            &= \argmax_{ \dtree \in \calH } \;
                         \sum_{i, \y}  \biggr[
                              \sum_{p \in {\cal N}, \atop y^p=c } \underbrace{ \lambda_{ ( i, \y ) } \dtree_{y^p}^{(1)}(\x_i^{p})}_\text{positive}
                              \\ \notag
                             & -
                            \sum_{p \in {\cal N},\atop  y_i^p=c } \underbrace{\lambda_{ ( i, \y ) } \dtree_{y_i^p}^{(1)}(\x_i^{p})}_\text{negative} \biggr].
       \end{align}
        To solve the above optimization problems, we here train $K$ weighted decision tree classifiers.
        Specifically, when training decision trees for the $c$-th class, the training data is composed of those superpixels whose ground truth label or predicted label is equal to the category label $c$.
        Since the output of the decision tree is in $\{0, 1 \}$ and  $\lambda_{(i, \y)} \geq 0$, the maximization in Eqn. \eqref{eq:dtree1} is achieved if $\dtree_c^{(1)}$ outputs 1 for each of the superpixel $p$ with $y^p=c$, and outputs 0 for each of the superpixel $p$  with $y_i^p=c$.
        Therefore, as indicated by the horizontal curly braces in Eqn. \eqref{eq:dtree1},  superpixels with the predicted labels of category $c$ are used as positive training examples, while superpixels with ground truth labels of category c are used as negative training examples. The dual solution $\blambda$ serves as weightings of the training data.

        For the pairwise part, we generate one decision tree in each column generation iteration, hence $\dt^{(2)\star}$ can be written as $\dt^{(2)\star}= [\dtree^{(2)\star}]$, the new decision tree for the pairwise part is generated as:
        \begin{align}  \label{eq:dtree2}
           \dtree^{(2)\star}(\cdot, \cdot) &= \argmax_{ \dtree \in \calH } \;
                         \sum_{i, \y  }  \lambda_{ ( i, \y ) }
                          \biggr[  \sum_{(p,q) \in {\cal S}} \underbrace{ \dtree^{(2)}(\x^{p}, \x^{q}) I (y^{p} \neq y^{q} ) }_\text{positive} \notag \\
                          &-  \sum_{(p,q) \in {\cal S}} \underbrace{\dtree^{(2)}(\x^{p}, \x^{q}) I (y_i^{p} \neq y_i^{q} )}_\text{negative} \biggr].
        \end{align}
        Similar to the unary case, we train a weighted decision tree classifier with $\blambda$ as training example weightings.
		The positive and negative training data are indicated by the horizontal curly braces in Eqn. \eqref{eq:dtree2}.
        $\dtree^{(2)}$ is the response of a decision tree applied on the pairwise features constructed by two neighbouring superpixels ($\x^p$, $\x^q$), \eg, color differences or shared boundary lengths.

        With the above analysis, we can now apply column generation to jointly learn the decision trees $\dt^{(1)}, \dt^{(2)}$ and $\w$.
        The column generation (CG) procedure iterates the following two steps:

        1) Solve Eqn. \eqref{eq:dtree1}, Eqn. \eqref{eq:dtree2} to generate decision trees $\dt^{(1)\star}$, $\dt^{(2)\star}$;

        2) Add $\dt^{(1)\star}$ and $\dt^{(2)\star}$ to working set $\mathcal W_{\dt}$ and resolve for the primal solution $  \w $ and dual solution $\blambda$.

\begin{figure*}[t]
    \centering
    \includegraphics[width=.164\linewidth, height=0.6548in]{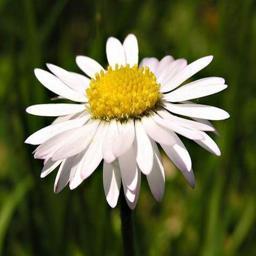}
    \includegraphics[width=.164\linewidth, height=0.6548in]{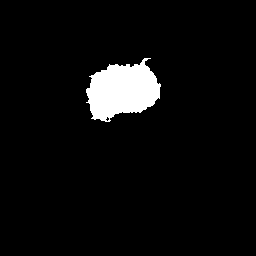}
    \includegraphics[width=.164\linewidth, height=0.6548in]{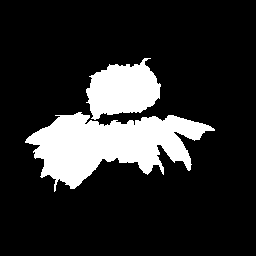}
    \includegraphics[width=.164\linewidth, height=0.6548in]{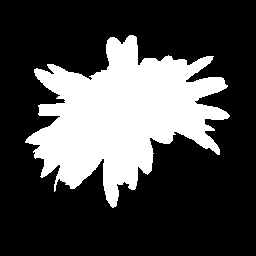}
    \includegraphics[width=.164\linewidth, height=0.6548in]{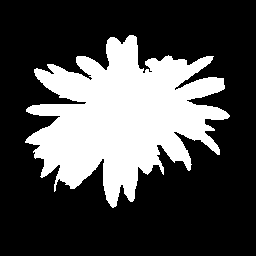}   \\

\vspace{.1cm}

    \includegraphics[width=.164\linewidth, height=0.6548in]{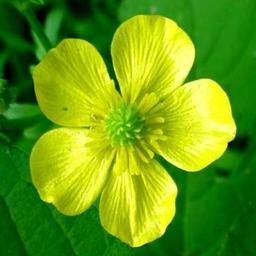}
    \includegraphics[width=.164\linewidth, height=0.6548in]{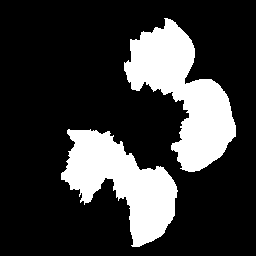}
    \includegraphics[width=.164\linewidth, height=0.6548in]{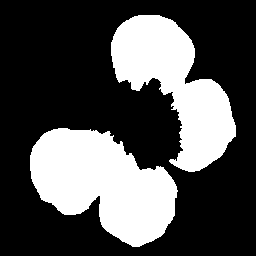}
    \includegraphics[width=.164\linewidth, height=0.6548in]{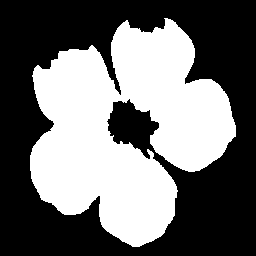}
    \includegraphics[width=.164\linewidth, height=0.6548in]{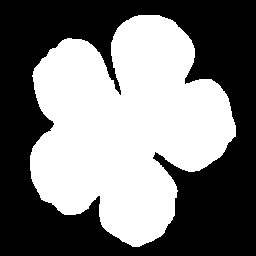}   \\
    \caption{Segmentation examples (each row being an example) produced by our model on images from the Oxford 17 Flower dataset with
    different column generation iterations.
        From left to right: Test images,
        2nd, 4th, 6th, and 10th iteration.
    }
    \label{fig:seg_flower}
     \end{figure*}

		We show two segmentation examples on the Oxford flower dataset produced by our method with different CG iterations in Fig. \ref{fig:seg_flower}. As can be seen, our method refines the segmentation with the increase of CG iterations. Since this dataset is relatively simple, a few CG iterations are enough to get satisfactory results.

		For solving the primal problem in the second step,
        it involves
        a large number of constraints due to the large output space $ \{ \y \in \calY \} $. We next show how to apply the cutting-plane technique \cite{Joachims06} to efficiently solve this problem.

\subsection{Speeding up optimization using cutting-plane}
\label{sec:one_slack}
To apply cutting-plane for solving the optimization in Eqn. \eqref{eq:ssvm_dt}, we first derive its \oneslack formulation.
The \oneslack\  \structsvm formulation was first introduced by \cite{Joachims06}.
The \oneslack formulation of our method can be written as:
    \begin{align}
        \label{eq:oneslack}
      \min_{\w \geq 0, \xi \geq 0}  \;\;
      &
      \tfrac{1}{2} \fnorm{\bw}^2 + C \xi \notag \\
      \st  \;\; &
       \frac{1}{m} \w^ \T
      \biggl[
                 \sum_{i=1}^m r_i
                \cdot
                        \left[ \Psi(\y, \x_i; \dt) - \Psi(\y_i, \x_i; \dt) \right]
        \biggr]    \notag \\
                &  \geq
                 \frac{1}{ m }  \sum_{i=1}^m r_i\loss(\y_i,\y ) - \xi,
                \forall {\boldsymbol r} \in\{0,1\}^m;
      \forall \y \in \calY.
    \end{align}

    Cutting-plane methods work by finding the most violated constraint for each example $i$
    \begin{align} \label{eq:inf_crf}
    \y_i^{\star} = \argmin  \w^{\T} \Psi(\y, \x; \dt) - \loss ( \y_i, \y)
    \end{align}
    at every iteration and add it to the constraint working set.
    The sketch of our method is summarized in Algorithm \ref{ALG:alg_structens},
 which calls Algorithm \ref{ALG:alg_cuttingPlane} to solve the \oneslack optimization.

\begin{algorithm}[t]
\caption{\footnotesize \structens using column generation}
\footnotesize{
1. {\bf Input:} training examples $ (\x_1; \y_1), (\x_2; \y_2) ,\cdots
    $; maximum iteration number.

    2. {\bf Initialize}
    ($\blambda$, $\y$), and decision tree working set $\mathcal W_{\dt} \leftarrow \emptyset $

    3. {\bf Repeat}

    4.$\quad-$ Find decision trees $\dt^{\star}$ by solving Eqn. \eqref{eq:dtree1}, Eqn. \eqref{eq:dtree2}.
     Add $\dt^{\star}$ to working set $\mathcal W_{\dt}$.

    5.$\quad-$
    Call Alg. \ref{ALG:alg_cuttingPlane} using working set $\mathcal W_{\dt}$ to solve for
    $ \w$ and $\blambda$.

    6. {\bf Until}
	the maximum iteration is reached.

    7. {\bf Output:}
    $\w $, $\dt \in \mathcal W_{\dt}$.
}
\label{ALG:alg_structens}
\end{algorithm}

\begin{algorithm}[t]
\caption{\footnotesize Cutting-planes for solving the \oneslack primal}
\footnotesize{

1: {\bf Input:} cutting-plane termination threshold $\epsilon_{\rm cp}
$,
and inputs from Alg. \ref{ALG:alg_structens}.

2: {\bf Initialize:} working set $\mathcal{W}\leftarrow \emptyset
$;  $\boldsymbol r$.

    3: {\bf Repeat}

    4: $\quad-$ $\mathcal{W}\leftarrow \mathcal{W} \cup \{ (r_1,
    \dots, r_m, \y_1^{\star},\dots,\y_m^{\star})  \} $.

    5: $\quad-$ Obtain primal and dual solutions $\w,\xi$; $\blambda $
    by solving \eqref{eq:oneslack} on $\mathcal{W}$.

    6: $\quad-$ {\bf For} $i=1, \dots, m$

	7: \; \; \; \;  \; \; \; Solve the inference problem in Eqn. \eqref{eq:inf_crf} using Graph-Cut to find the most violated $\y_i^{\star}$.

    8: $\quad\;\;\,$ {\bf End for}

	9: {\bf Until}  \\
$
      \frac{1}{m} \w^ \T
  \biggl[
             \sum\limits_{i=1}^m r_i
                    \left[ \Psi(\y_i^{\star}, \x_i) - \Psi(\y_i, \x_i) \right]
    \biggr]
            \geq
            \frac{1}{ m } \sum\limits_{i=1}^m r_i\loss(\y_i,\y_i' ) - \xi -
            \epsilon_{\rm cp}.
$

	10: {\bf Output:} $\w, \blambda$.

}
\label{ALG:alg_cuttingPlane}
\end{algorithm}

\paragraph{Implementation details}
To deal with the unbalanced appearance of different categories in the dataset, we define $\loss ( \y_i, \y)$ as weighted Hamming loss, which weighs errors for a given class inversely proportional to the frequency it appears in the training data, as similar in \cite{Lucchi12}.
In the inference problem of Eqn. \eqref{eq:inf_crf},
when using the hamming loss as the label cost $\loss$, the label cost term can be absorbed into the unary part.
We therefore can apply Graph-cut to efficiently solve Eqn. \eqref{eq:inf_crf}.
As for more complicated label cost functions, an efficient inference algorithm is proposed in \cite{Bauer2016}.
During each CG iteration, our method first solves Eqn. \eqref{eq:dtree1}, \eqref{eq:dtree2} given the current $\x$ and $\xi$, and then solves a quadratic programming (QP) problem given $\dt$.
When solving Eqn. \eqref{eq:dtree1}, \eqref{eq:dtree2}, we train weighted decision tree classifiers using the highly optimized decision tree training method of \cite{Piotr13}.

  \paragraph{Discussions on the submodularity}
It is known that if graph cuts are to be applied to achieve globally optimum labelling in segmentation, the energy function must be submodular.
For foreground/background segmentation in which a (super-)pixel label takes value in $\{0, 1\}$,
we show that our method keeps this submodular property.
It is commonly known that an energy function is submodular if its pairwise term satisfies: $\eta_{pq}(0, 0) + \eta_{pq}(1, 1) \leq \eta_{pq}(0, 1) + \eta_{pq}(1, 0)$.  Recall that our pairwise energy is written as $\eta_{pq}(y^{p}, y^{q}) = \w^{(2)\T} \dt^{(2)}(\x^{p}, \x^{q}) I (y^{p} \neq y^{q} ) $.
Clearly we have ($\eta_{pq}(0, 0)=\eta_{pq}(1, 1)=0$) because of the indicator function  $I(y^p \neq y^q)$. The second thing is to ensure $\eta_{pq}(1, 0)  + \eta_{pq}(0, 1) \geq 0$. Given the non-negativeness constraint we impose on $\w$ in our model, and the output of decision trees in our method taking values from $\{0, 1\}$, we have $\eta_{pq}(1, 0)\geq 0$ and  $\eta_{pq}(0, 1) \geq 0$. We thus accomplish the proof of the submodularity of our model. In the case of multi-object segmentation, the inference is done by the  $\alpha$-expansion of graph cuts.

 \paragraph{Discussions on the non-negative constraint on $\w$}
Our learning framework aligns with boosting methods, where we learn a non-negative weighted ensemble of weak structured learners (constructed by decision trees), which is analogous to weak learners in boosting methods. This is similar to boosting methods, such as AdaBoost, LPBoost \cite{lpboost}, where the non-negative weighting is commonly used.
Further, a weak structured learner generated by our column generation method is expected to make positive contribution to the learning objective. If it is of no use to the objective, the weight will approach zero.
Therefore it is reasonable to enforce the non-negative constraint on $\w$.

\section{Experiments}
To demonstrate the effectiveness of the proposed method, we first compare our model with some most related baseline methods, which are \svm, \adaboost and \ssvm.
In section \ref{sec: comp2}, we show that our method achieves state-of-the-art results by exploiting recent advances in feature learning  \cite{Coates11, Alex12}.

\subsection{Experimental setup}
The datasets evaluated here include three binary datasets (Weizmann horse, Oxford flower and Graz-02) and two multi-class datasets (MSRC-21 and PASCAL VOC 2012). %
The Weizmann horse dataset\footnote{\url{ http://www.msri.org/people/members/eranb/} }  consists of 328 horse images from various backgrounds, with groundtruth masks available for each image. We use the same data split as in \cite{Bertelli11} and \cite{Kuettel12}.
The Oxford 17 category flower dataset \cite{Flower17} is composed of 849 flower images.
Those with too small foreground are removed, which leaves 753 for segmentation purpose \cite{Flower17}. The data split stated in \cite{Flower17} is used to perform the evaluation. During our experiment, images of the Weizmann horse and the Oxford flower datasets are resized to 256$\times$256.
The Graz-02
dataset\footnote{\url{http://www.emt.tugraz.at/~pinz/} } contains 3 categories (bike, car and people).
This dataset is considered challenging as the objects appear at various background and with different poses.
We follow the evaluation protocol in \cite{Marszalek07} to use 150 for training and 150 for testing for each category.
The MSRC-21 dataset \cite{Shotton08} is a popular multi-class segmentation benchmark with 591 images containing objects from 21 categories. We follow the standard split to divide the dataset into training/validation/test subsets.
The PASCAL VOC 2012 dataset \footnote{\url{http://host.robots.ox.ac.uk/pascal/VOC/voc2012/} } is a widely used benchmark for semantic segmentation, which contains 2913 images from the trainval set and 1456 images from the test set, making up 21 categories. Unlike many state-of-the-arts methods such as \cite{crfasrnn_iccv2015}, we do not use any additional training data for this dataset.

We start with over-segmenting the images into superpixels using SLIC \cite{slic}, with $\sim$ 700 superpixels generated per image.
We extract dense SIFT descriptors and color histograms around each superpixel centroid with different block sizes (12$\times$12, 24$\times$24, 36$\times$36). The dense SIFT descriptors are then quantized into bag-of-words features using nearest neighbour search with a codebook size of 400.
We construct four types of pairwise features also using different block sizes to enforce spatial smoothness, which are color difference in LUV space, color histogram difference, texture difference in terms of LBP operators as well as shared boundary length \cite{Fulkerson09}.
The column generation iteration number of our \sens is set to 50 based on a validation set.
We learn tree potentials with the tree depth being $2$.
Training on the MSRC-21 dataset on a standard PC machine takes around 16 hours.

\subsection{Comparing with baseline methods} \label{sec:compBaseline}
We first compare \sens with some conventional methods, which are linear \svm, \adaboost and \ssvm to demonstrate the superiority of our method.
For \svm and \adaboost, each superpixel is classified independently without \crf.
We mainly evaluate on the more challenging Graz-02 and MSRC-21 dataset in this part.
The regularization parameter C of \svm, \ssvm and our \sens are selected from $\{1, 10, 100, 1000 \}$ based on a validation set. We use depth-2 decision trees for training AdaBoost and our \sens. The maximum iteration number of \adaboost is chosen from $\{$50, 100, 200$\}$.
For our method, we treat the foreground and background as two categories in the binary case to learn class-wise potentials.

\paragraph{Graz-02} For a comprehensive evaluation, we use two  measurements to quantify the performance on the Graz-02 dataset, which are intersection over union score and the pixel accuracy (including foreground and background). We report the results in Table~\ref{tab:seg_graz02_baseline}. As can be observed, \adaboost based on a depth-2 decision tree performs better than the linear \svm. On the other hand, structured methods which jointly consider local information and spatial consistency are able to significantly outperform  the simple binary models. By introducing nonlinear and class-wise potential learning,  our method is able to gain further improvement over \ssvm.

\paragraph{MSRC-21}
We learn class-wise potentials using our \sens for each of the 21 classes on the MSRC dataset.
The compared results are summarized in Table~\ref{tab:seg_msrc} (upper part). Similar conclusions can be drawn as on the Graz-02 dataset and our \sens again outperforms all its baseline competitors.

\begin{table} [t]
\centering
\resizebox{.96\linewidth}{!}
  {
  \begin{tabular}{| r | c| c| c |}
    \hline
    Category &bike &car &people \\
  \hline
    &\multicolumn{3}{c|}{intersection/union (foreground, background)($\%$)}  \\
  \hline
  \svm  &67.8 (51.9, 83.8) &69.7 (46.8, 92.6)   &65.0 (44.5, 85.5) \\
  \adaboost  &71.2 (57.6, 84.9)    &71.0 (49.4, 92.6)  &67.7 (48.7, 86.7) \\
  \ssvm &72.2 (58.6, 85.8)  &76.9 (60.0, 94.2)  &70.9 (53.8, 87.9) \\
   \sens &76.4 (65.0, 87.8) &79.5 (64.0, 95.0) &74.2 (58.7, 89.7)  \\
   \sens (FL)  &\textbf{78.3} (67.7, 88.9) &\textbf{83.0} (70.1, 95.9)  &\textbf{75.7} (61.0, 90.5)  \\
   \hline \hline
     &\multicolumn{3}{c|}{pixel accuracy (foreground, background)($\%$)}  \\
   \hline
   \svm  &79.5 (67.4, 91.5)  &77.3 (57.2, 97.3)   &77.7 (63.8, 91.6) \\
  \adaboost  &83.8 (77.3, 90.3)  &80.1 (63.5, 96.6) &80.5 (69.0, 91.9) \\
  \ssvm  &83.8 (76.1, 91.6)  &85.5 (73.8, 97.2)  &83.9 (75.8, 92.1) \\
   \sens &87.8 (83.9, 91.8) &87.0 (76.4, 97.7)  &85.9 (78.4, 93.4)  \\
   \sens (FL) &\textbf{89.1} (85.8, 92.4)  &\textbf{90.0} (82.1, 98.0)  &\textbf{86.9} (80.0, 94.0) \\
   \hline
  \end{tabular}
  }
\caption{The average intersection-over-union score and average pixel accuracy comparison on the Graz-02 dataset. We include the foreground and background results in the brackets. Our method \sens with nonlinear and class-wise potentials learning performs better than all the baseline methods.}
\label{tab:seg_graz02_baseline}
\end{table}

\begin{table*} [t]
\centering
\resizebox{0.95\linewidth}{!}
{
\begin{tabular}{ c c c c c c c c c c c c c c c c c c c c c c |c c }
&\rotatebox{90}{building}  &\rotatebox{90}{grass}  &\rotatebox{90}{tree}  &\rotatebox{90}{cow}  &\rotatebox{90}{sheep}  &\rotatebox{90}{sky}  &\rotatebox{90}{aeroplane}  &\rotatebox{90}{water}  &\rotatebox{90}{face}  &\rotatebox{90}{car}  &\rotatebox{90}{bicycle}  &\rotatebox{90}{flower} &\rotatebox{90}{sign}  &\rotatebox{90}{bird}  &\rotatebox{90}{book}  &\rotatebox{90}{chair}  &\rotatebox{90}{road}  &\rotatebox{90}{cat}  &\rotatebox{90}{dog}  &\rotatebox{90}{body}  &\rotatebox{90}{boat}  &\rotatebox{90}{\bf{Average}} &\rotatebox{90}{\bf{Global}} \\
\hline  \hline
\svm  &54  &92  &73 & 41  &54  &80 &51 &67 &51 &41 &59 &41 &28 &8  &64  &17  &75  &41  &23  &20   &7  &47.0  &63.7  \\
\adaboost   &68    &92    &83    &48    &58    &87    &43    &69    &58    &43    &64    &41    &32    &14    &70   &28    &79    &47    &22    &41    &6   &52.0  &68.6  \\
\ssvm   &65    &92    &81    &42    &76    &84    &65    &70    &75    &54    &87    &62    &31    &14    &76    &31    &78    &61    &30   &25    &2  &57.2  &70.8\\
\sens  &53    &87    &85    &59    &84    &90    &77    &82    &81    &54    &90    &57    &62    &22    &81    &59    &80    &71    &26    &49    &15   &\textbf{64.9}   &\textbf{73.9}  \\
\hline   \hline
\sens (FL)      &66    &95    &89    &83    &89    &90    &90    &83    &76    &74    &83    &71    &69    &46    &87    &73    &87    &84    &53    &68    &20  &75.1  &82.2  \\
\sens (CNN)   &73    &96    &89  &82  &92    &96   &89    &86    &93   & 78    &86    &91    &71    &75    &85    &76    &86    &91    &63   & 83    &41  &\textbf{82.0}  &\textbf{86.2}\\
Shotton \etal \cite{Shotton08} &49 &88 &79 &97 &97 &78 &82 &54 &87  &74 &72 &74 &36  &24  &93  &51 &78  &75 &35 &66 &18  &67  &72 \\
Ladicky \etal \cite{Ladicky09}  &80 &96 &86 &74 &87 &99 &74 &87 &86 &87 &82 &97 &95 &30 &86 &31 &95 &51 &69 &66 &9  &75  &86 \\
Gonfaus \etal \cite{Harmony10} &60 &78 &77 &91 &68 &88 &87 &76 &73  &77 &93 &97 &73  &57  &95  &81 &76 &81  &46 &56 &46 &75 &77 \\
Lucchi \etal \cite{Lucchi12} &59 &90 &92 &82 &83 &94 &91 &80 &85 &88 &96 &89 &73 &48 &96 &62 &81 &87 &33 &44 &30 &76 &82 \\
Lucchi \etal \cite{Lucchi13}  &67 &89 &85 &93 &79 &93 &84 &75 &79  &87 &89 &92 &71  &46  &96  &79 &86  &76 &64 &77 &50  &78.9  &83.7  \\
\hline
\end{tabular}
}
\caption{Segmentation results on the MSRC dataset. We report the pixel-wise accuracy for each category as well as the average per-category scores and the global pixel-wise accuracy. (1) The upper part presents the comparison with baseline methods, which all use bag-of-words and color histogram features. Our method \sens gains impressive improvements over \ssvm while far better than simple linear models. (2) The lower part shows the results of our method using unsupervised feature learning and CNN features  (denoted as \sens (FL) and \sens (CNN) respectively) compared with state-of-the-art methods on this dataset.
}
\label{tab:seg_msrc}
\end{table*}

\begin{figure*} [t]
\centering
    \includegraphics[width=.1\linewidth, height=0.45in]{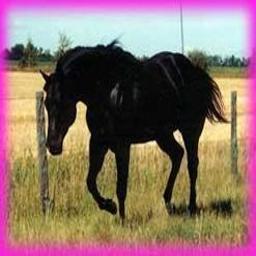}
     \includegraphics[width=.1\linewidth, height=0.45in]{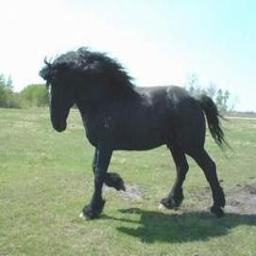}
     \includegraphics[width=.1\linewidth, height=0.45in]{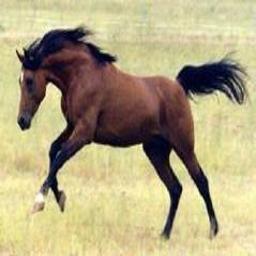}
	\includegraphics[width=.1\linewidth, height=0.45in]{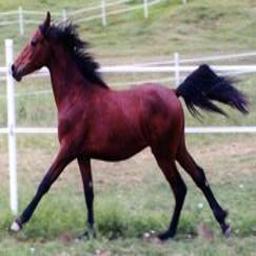}
    \includegraphics[width=.1\linewidth, height=0.45in]{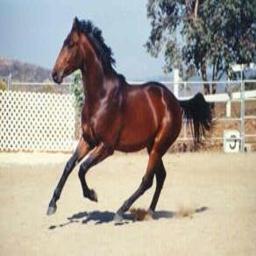}
     \includegraphics[width=.1\linewidth, height=0.45in]{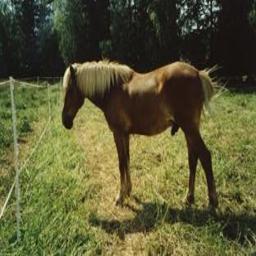}
     \includegraphics[width=.1\linewidth, height=0.45in]{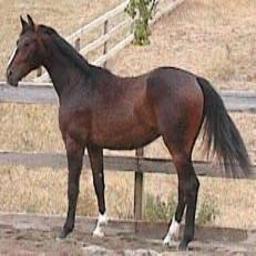}
     \includegraphics[width=.1\linewidth, height=0.45in]{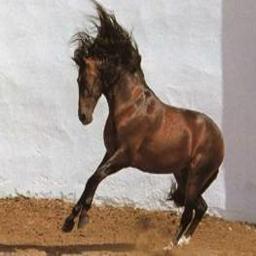} \\

     \vspace{.1cm}

     \includegraphics[width=.1\linewidth, height=0.45in]{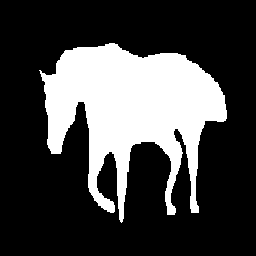}
     \includegraphics[width=.1\linewidth, height=0.45in]{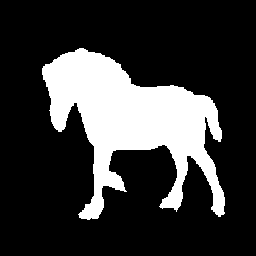}
     \includegraphics[width=.1\linewidth, height=0.45in]{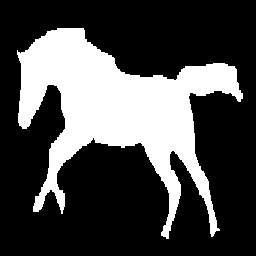}
	\includegraphics[width=.1\linewidth, height=0.45in]{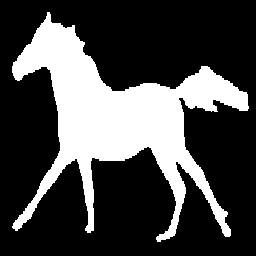}
    \includegraphics[width=.1\linewidth, height=0.45in]{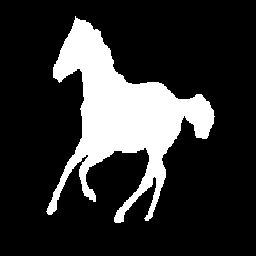}
     \includegraphics[width=.1\linewidth, height=0.45in]{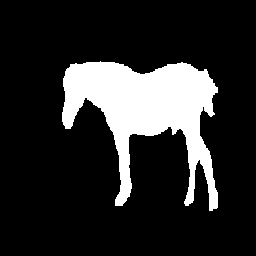}
     \includegraphics[width=.1\linewidth, height=0.45in]{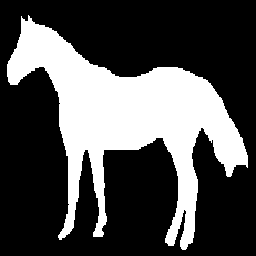}
     \includegraphics[width=.1\linewidth, height=0.45in]{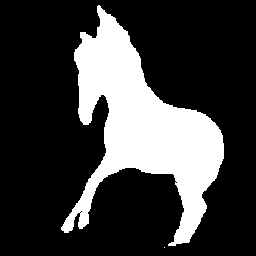} \\

\vspace{.1cm}

     \includegraphics[width=.1\linewidth, height=0.45in]{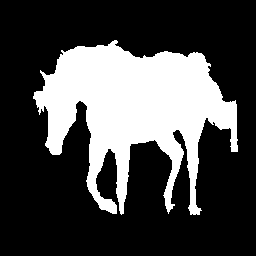}
     \includegraphics[width=.1\linewidth, height=0.45in]{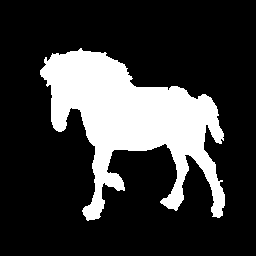}
     \includegraphics[width=.1\linewidth, height=0.45in]{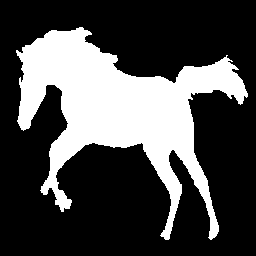}
	\includegraphics[width=.1\linewidth, height=0.45in]{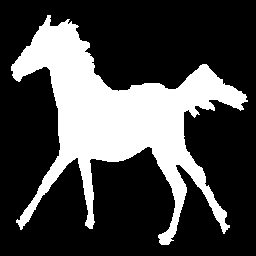}
    \includegraphics[width=.1\linewidth, height=0.45in]{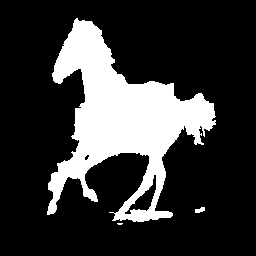}
     \includegraphics[width=.1\linewidth, height=0.45in]{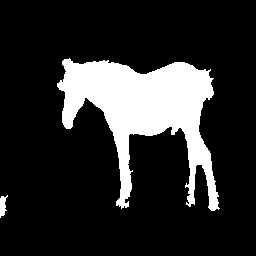}
     \includegraphics[width=.1\linewidth, height=0.45in]{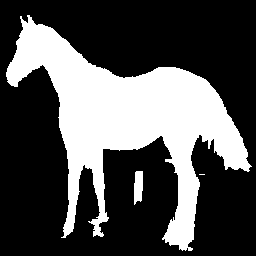}
     \includegraphics[width=.1\linewidth, height=0.45in]{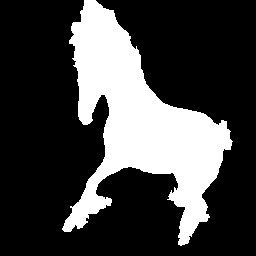} \\

\caption{Segmentation examples on Weizmann horse. 1st row: Test images; 2nd row: Ground truth; 3rd row: segmentation produced by \sens. }
    \label{fig:seg_horse}

\end{figure*}

\begin{figure*} [!t]
\centering
    \includegraphics[width=.11\linewidth, height=0.45in]{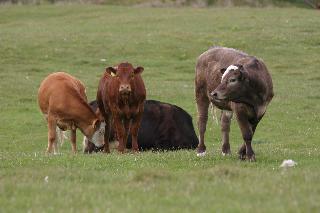}
     \includegraphics[width=.113\linewidth, height=0.45in]{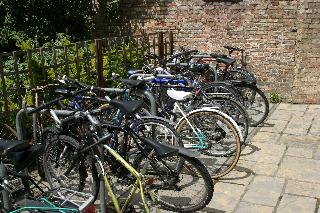}
	\includegraphics[width=.11\linewidth, height=0.45in]{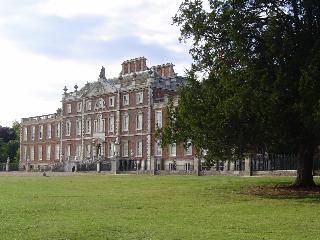}
    \includegraphics[width=.11\linewidth, height=0.45in]{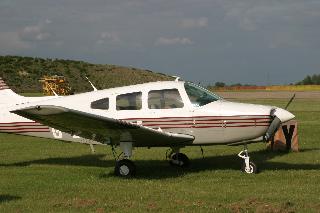}
     \includegraphics[width=.11\linewidth, height=0.45in]{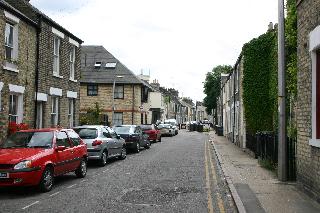}
     \includegraphics[width=.11\linewidth, height=0.45in]{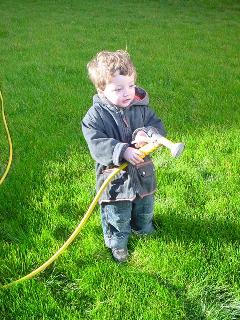}
     \includegraphics[width=.11\linewidth, height=0.45in]{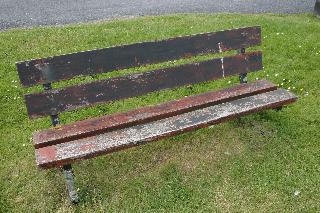}
     \includegraphics[width=.11\linewidth, height=0.45in]{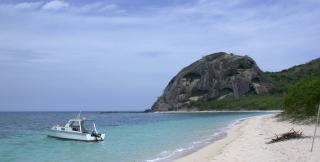}\\

     \vspace{.1cm}

     \includegraphics[width=.11\linewidth, height=0.45in]{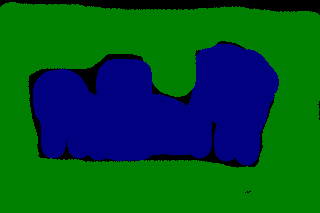}
     \includegraphics[width=.11\linewidth, height=0.45in]{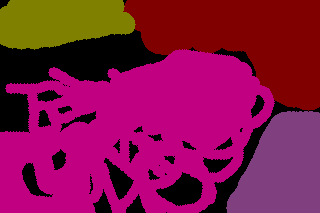}
	\includegraphics[width=.11\linewidth, height=0.45in]{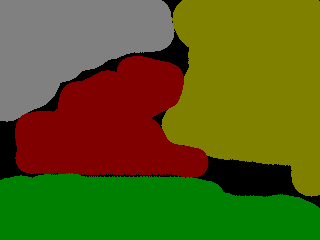}
    \includegraphics[width=.11\linewidth, height=0.45in]{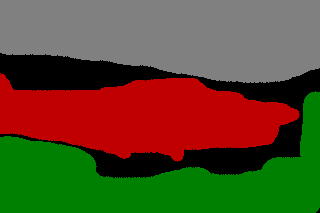}
     \includegraphics[width=.11\linewidth, height=0.45in]{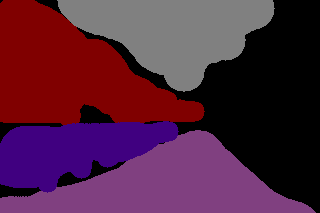}
     \includegraphics[width=.11\linewidth, height=0.45in]{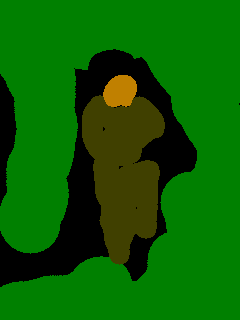}
     \includegraphics[width=.11\linewidth, height=0.45in]{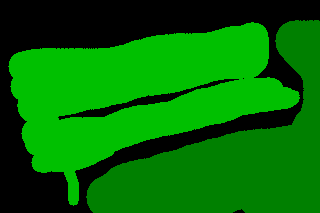}
     \includegraphics[width=.11\linewidth, height=0.45in]{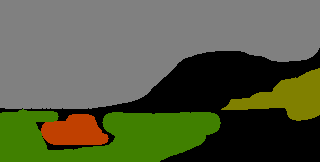}\\

     \vspace{.1cm}

     \includegraphics[width=.11\linewidth, height=0.45in]{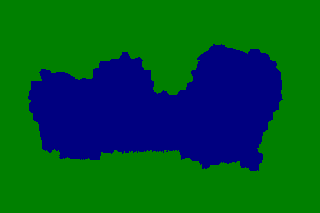}
     \includegraphics[width=.11\linewidth, height=0.45in]{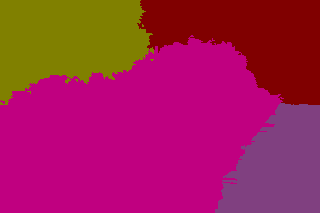}
	\includegraphics[width=.11\linewidth, height=0.45in]{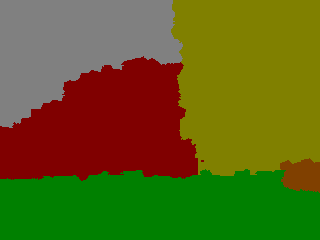}
    \includegraphics[width=.11\linewidth, height=0.45in]{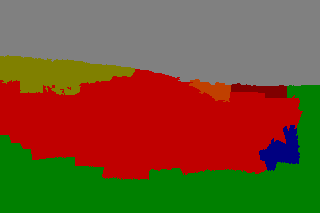}
     \includegraphics[width=.11\linewidth, height=0.45in]{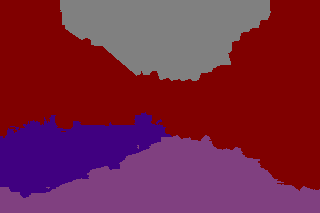}
     \includegraphics[width=.11\linewidth, height=0.45in]{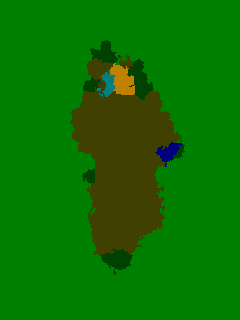}
     \includegraphics[width=.11\linewidth, height=0.45in]{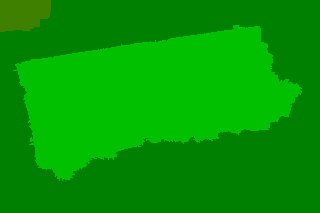}
     \includegraphics[width=.11\linewidth, height=0.45in]{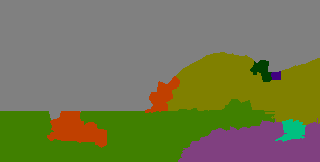}\\

\caption{Segmentation examples on MSRC. 1st row: Test images; 2nd row: Ground truth; 3rd row: \sens with unsupervised feature learning. }

    \label{fig:seg_msrc}

\end{figure*}

\begin{figure*} [!t]
\centering
    \includegraphics[width=.11\linewidth, height=0.45in]{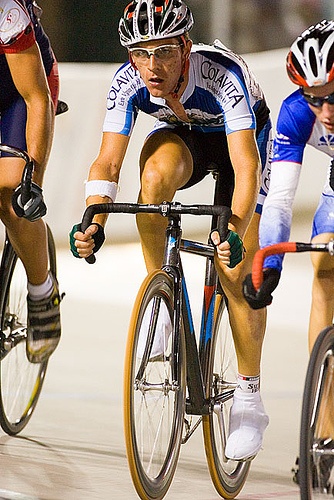}
     \includegraphics[width=.113\linewidth, height=0.45in]{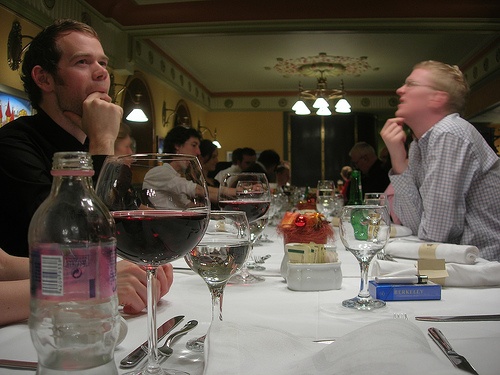}
  \includegraphics[width=.11\linewidth, height=0.45in]{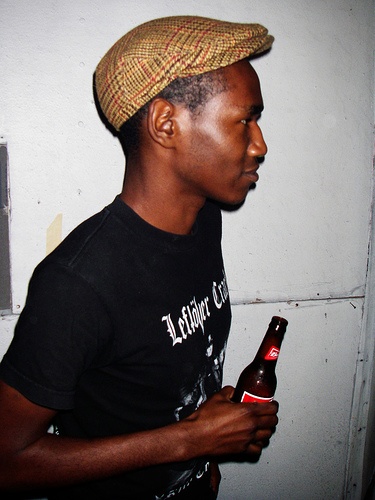}
    \includegraphics[width=.11\linewidth, height=0.45in]{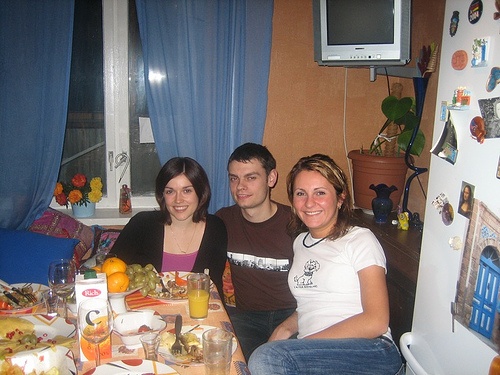}
     \includegraphics[width=.11\linewidth, height=0.45in]{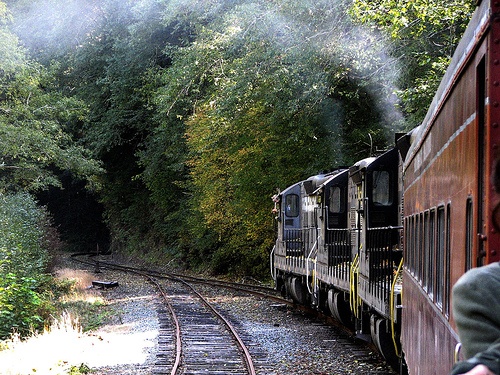}
     \includegraphics[width=.11\linewidth, height=0.45in]{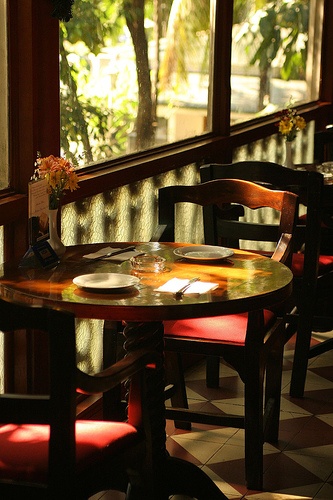}
     \includegraphics[width=.11\linewidth, height=0.45in]{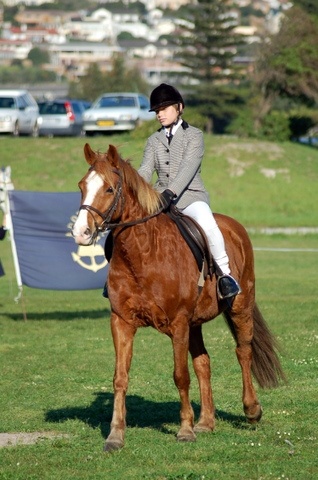}
     \includegraphics[width=.11\linewidth, height=0.45in]{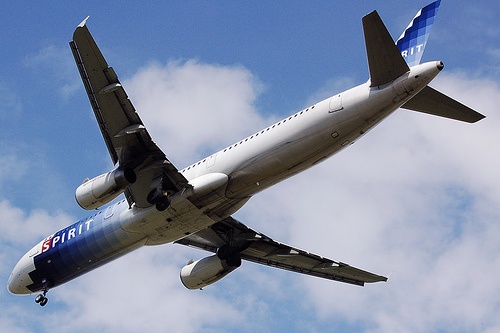}\\

     \vspace{.1cm}

     \includegraphics[width=.11\linewidth, height=0.45in]{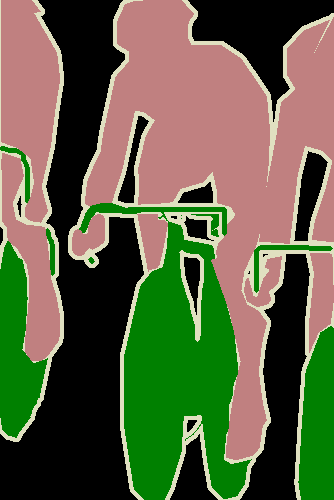}
     \includegraphics[width=.11\linewidth, height=0.45in]{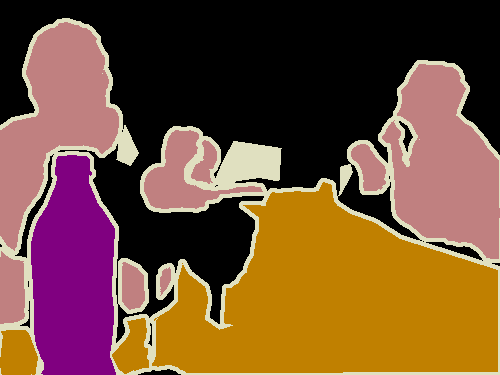}
  \includegraphics[width=.11\linewidth, height=0.45in]{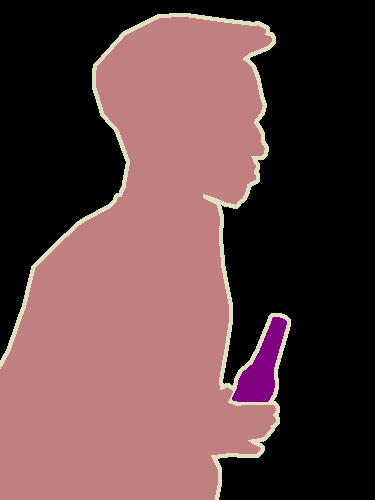}
    \includegraphics[width=.11\linewidth, height=0.45in]{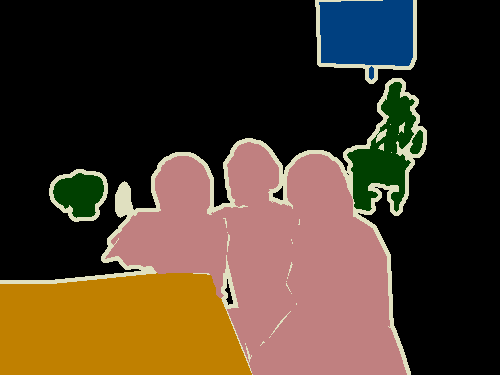}
     \includegraphics[width=.11\linewidth, height=0.45in]{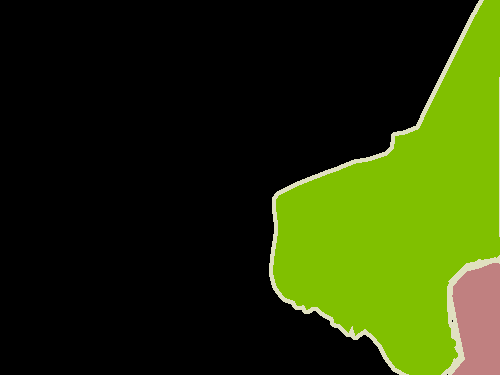}
     \includegraphics[width=.11\linewidth, height=0.45in]{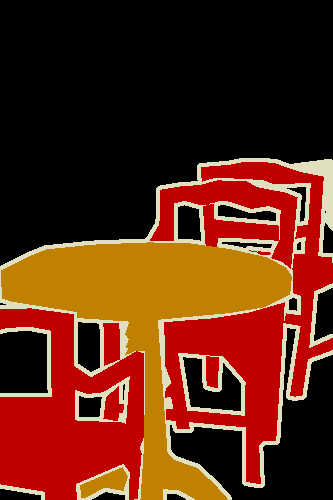}
     \includegraphics[width=.11\linewidth, height=0.45in]{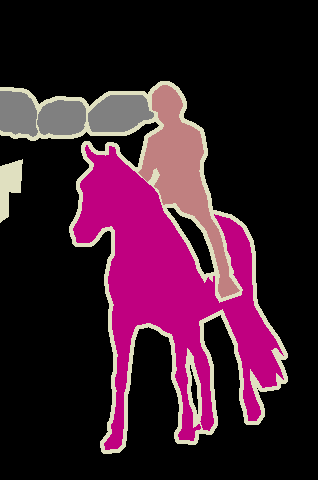}
     \includegraphics[width=.11\linewidth, height=0.45in]{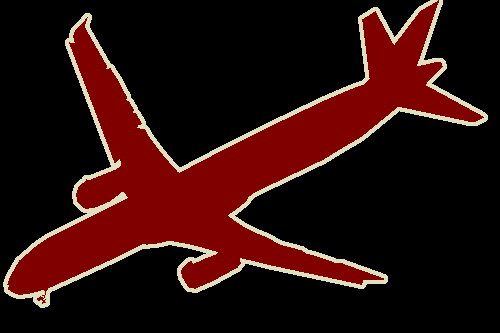}\\

\vspace{.1cm}

     \includegraphics[width=.11\linewidth, height=0.45in]{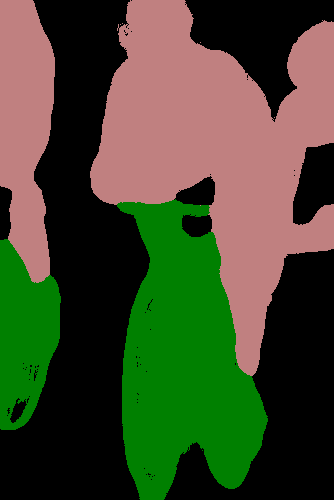}
     \includegraphics[width=.11\linewidth, height=0.45in]{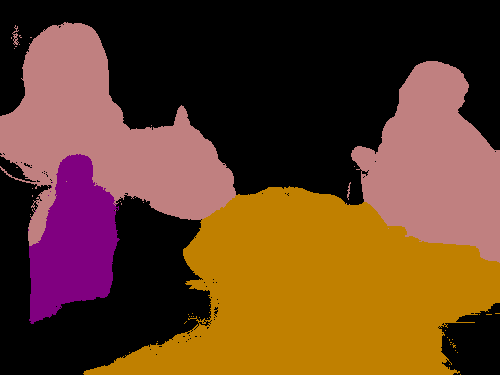}
  \includegraphics[width=.11\linewidth, height=0.45in]{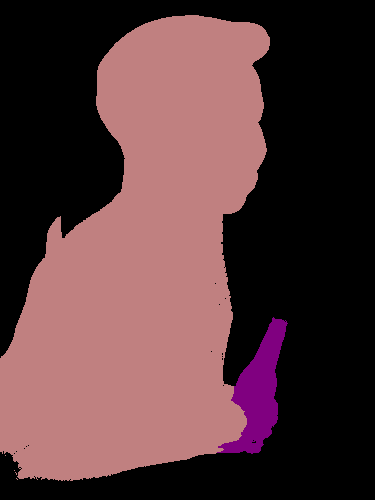}
    \includegraphics[width=.11\linewidth, height=0.45in]{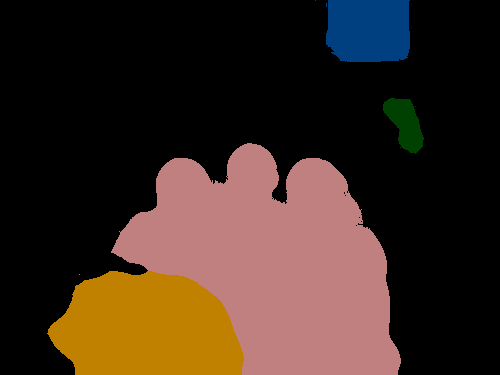}
     \includegraphics[width=.11\linewidth, height=0.45in]{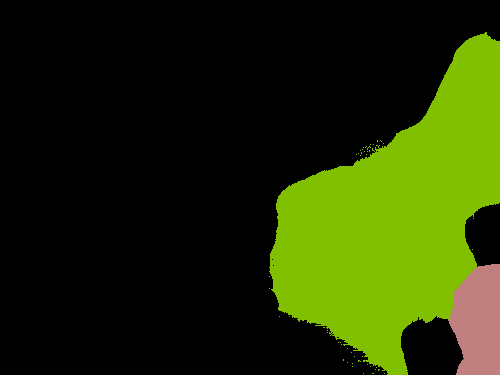}
     \includegraphics[width=.11\linewidth, height=0.45in]{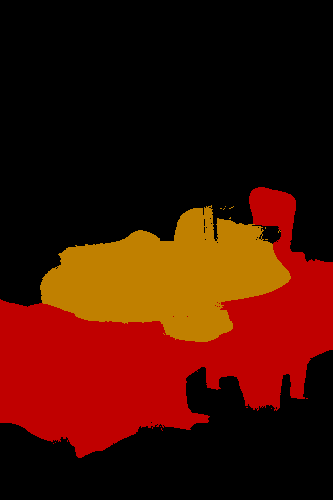}
     \includegraphics[width=.11\linewidth, height=0.45in]{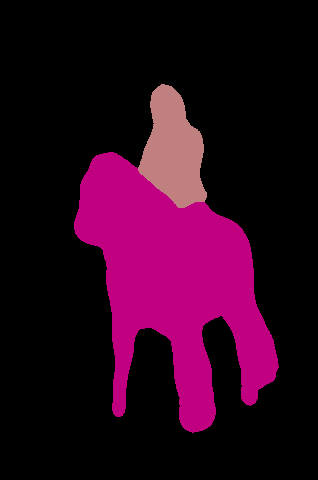}
     \includegraphics[width=.11\linewidth, height=0.45in]{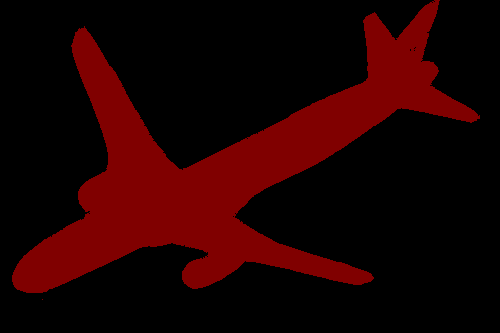}\\

\caption{Segmentation examples on PASCAL VOC 2012. 1st row: Images; 2nd row: Ground truth; 3rd row: \sens prediction results. }
    \label{fig:seg_voc}

\end{figure*}

\subsection{Comparing with state-of-the-art methods} \label{sec: comp2}
Since features play a pivotal role in the performance of vision algorithms,
 we exploit recent advances in feature learning to pursue state-of-the-art results, \ie, unsupervised feature learning \cite{Coates11} and convolutional neural networks (CNN) \cite{Alex12}.
Specifically, for the unsupervised feature learning, we first learn a dictionary $\bf B$ of size 400 and patch size 6$\times$6 based on the evaluated image dataset using Kmeans, and then use the soft threshold coding \cite{Coates11}
to encode patches extracted from each superpixel block. The final feature vectors (we call it encoding feature here) are obtained by performing a three-level max pooling over the superpixel block.
For the CNN features, we use the Alex model \cite{Alex12} trained on the ImageNet\footnote{\url{http://image-net.org}} to generate CNN features.
These two versions of our method are denoted as \sens (FL) and \sens (CNN) respectively.
We only report the results of \sens (CNN) on the MSRC-21 and PASCAL VOC 2012 datasets since our method already performs very well by using the encoding features on the three binary datasets.

\begin{table}
\centering
\resizebox{.53\linewidth}{!} {
\begin{tabular}{| l | c |  c | }
\hline
 Method   &Sa  &So \\
\hline
Levin \& Weiss \cite{Levin06} &\textbf{95.5} &-  \\
Cosegmentation \cite{cosegmentation10} &80.1 &- \\
Bertelli \etal \cite{Bertelli11}  & 94.6 &80.1 \\
Kuttel \etal \cite{Kuettel12} &94.7 &- \\
\hline
\sens (FL) &94.6	 &\textbf{80.4}	\\
\hline
\end{tabular}
}
\caption{Performance of different methods on the Weizmann Horse dataset. }
\label{tab:seg_horse}
\end{table}

\paragraph{Weizmann horse} We quantify the performance by the global pixel-wise accuracy $S_a$ and the foreground intersection over union score $S_o$, as did in \cite{Bertelli11}. $S_a$ measures the percentage of pixels correctly classified while $S_o$ directly reflects the segmentation quality of the foreground.
The results are reported in Table~\ref{tab:seg_horse}. Our method performs better than the kernel structural learning method of \cite{Bertelli11}, which may result from the fact that they only introduced nonlinearity into the unary part while our method achieves nonlinearity on both unary and pairwise terms.
The best $S_a$ score is obtained by \cite{Levin06}. However their method relies on an assumption that a perfect bounding box of the horse is available for each test image, which is not practically applicable. On the contrary, we provide a principal and general way of nonlinearly learning \crf parameters.
We show some segmentation examples of our method in Fig.~\ref{fig:seg_horse}.

\begin{table}
\centering
\resizebox{0.56\linewidth}{!} {
\begin{tabular}{| l | c |  c | }
\hline
 Method   &Sa  &So \\
\hline
Nilsback \etal \cite{Flower17} &- &94.0  \\
Bertelli \etal \cite{Bertelli11}  & 97.7 &92.3 \\
\hline
\sens (FL) &\textbf{98.0}	 &\textbf{94.2}	\\
\hline
\end{tabular}
}
\caption{Performance of different methods on the Oxford FLower dataset. Our method \sens performs better than the compared methods.}
\label{tab:seg_flower17}
\end{table}

\paragraph{Oxford flower}
As in \cite{Bertelli11}, we also use $S_a$ and $S_o$ to measure the performance on the Oxford flower dataset, and report the results in  Table~\ref{tab:seg_flower17}. Our method performs comparable to the original work of \cite{Flower17} on this dataset in terms of $S_o$ while again obtains better results  than the closely related state-of-the-art work of \cite{Bertelli11}. It is also worth noting that the method in \cite{Flower17} is very domain specific, which relies on modelling the flower's shape (center and petal), while ours is generally applicable.

\paragraph{Graz-02}
As in the work of \cite{Marszalek07}, \cite{Fulkerson08}, \cite{Aldavert10}, \cite{Kuettel12}, we also evaluate the F-score on the Graz-02 dataset besides the above mentioned intersection over union score and pixel accuracy.
The F-score is defined as $F=2pr/(p+r)$, where $p$ is the precision and $r$ is the recall. We summarize the results in Table~\ref{tab:seg_graz02} and Table~\ref{tab:seg_graz02_baseline}.
From Table~\ref{tab:seg_graz02}, it can be seen that our method significantly outperforms all the compared methods, which fully demonstrate the power of nonlinear and class-wise potential learning.
Furthermore, we can observe from Table~\ref{tab:seg_graz02_baseline} that compared with the previous results, adding more features help to  improve the performance.

\begin{table}
\centering
\resizebox{0.86\linewidth}{!} {
\begin{tabular}{| l | c |  c | c  | c |}
\hline
 Method   &bike  &car  &people  &average \\
\hline
Marszalek \& Schimid \cite{Marszalek07}  &61.8 &53.8 &44.1 & 53.2\\
Fulkerson \etal \cite{Fulkerson08} &66.4  &54.7 &51.4 &57.5 \\
Aldavert \etal \cite{Aldavert10} &71.9 &62.9 &58.6 &64.5  \\
Kuettel \etal \cite{Kuettel12} &63.2  &74.8 &66.4 &68.1\\
\hline
\sens (FL) &\textbf{80.7}	 &\textbf{82.4}	&\textbf{75.8}  &\textbf{79.5}
 \\
\hline
\end{tabular}
}
\caption{Comparing with state-of-the-art methods on the Graz-02 dataset. We report the F-score (\%) for each class and the average over classes. Our method \sens outperforms all the compared methods with a large margin.}
\label{tab:seg_graz02}
\end{table}

\paragraph{MSRC-21}
The compared results with state-of-the-art works are reported in the lower part of Table~\ref{tab:seg_msrc}.
As we can see, by incorporating more advanced features, our \sens gains significant improvements over the previous results which only use bag-of-words and color histogram features.
It is worth noting that our method performs better than the closely related work of Lucchi \etal \cite{Lucchi12} which claims exploiting non-linear kernels.
It has to be pointed out that we did not employ any global potentials (while in  \cite{Lucchi12}, they improve the global and average per-category accuracy from 70, 73 to 82 and 76 by adding global information). If global or higher potentials are incorporated into our model, further performance promotion can be expected.
We show some qualitative evaluation examples in Fig.~\ref{fig:seg_msrc}.

\paragraph{PASCAL VOC 2012}
We generate deep features of each superpixel by averaging the pixel-wise feature map scores within the superpixel obtained from a pretrained FCN model \cite{long_shelhamer_fcn}.
We then train our \sens model on the standard PASCAL VOC 2012 training dataset with the generated deep features.
Following the standard evaluation procedure for the Pascal VOC challenge, we upload our segmentation results to the test server and use the average intersection over union as the evaluation metric.
We compare against several state-of-the-art methods (\cite{Hariharan_cvpr2015}, \cite{Dai_cvpr2015}, \cite{long_shelhamer_fcn}, \cite{crfasrnn_iccv2015}) on the test set of the PASCAL VOC 2012 dataset.
The results are reported in Table~\ref{tab:seg_pascal_voc_2012}.
As seen from the table, our \sens beats the Hypercolum \cite{Hariharan_cvpr2015} and the CFM \cite{Dai_cvpr2015} and outperforms the FCN \cite{long_shelhamer_fcn} by a notable margin.
Although our method is triumphed by \cite{crfasrnn_iccv2015}, it should be noted that their result is obtained by using extra  training data (11,685 images vs 1456 images used for training our \sens).
Some qualitative evaluation examples of our method are illustrated in Fig.~\ref{fig:seg_voc}.

\begin{table}
\centering
\begin{tabular}{| l | c | }
\hline
 Method   &intersection/union  \\
\hline
CFM \cite{Dai_cvpr2015} &61.8 \\
Hypercolumn \cite{Hariharan_cvpr2015} &62.6 \\
FCN-8s \cite{long_shelhamer_fcn}  &62.2  \\
Zheng \etal \cite{crfasrnn_iccv2015}   &\textbf{72.0} \\
\hline
\sens   &65.4 \\
\hline
\end{tabular}

\caption{The average intersection-over-union scores of different methods on the PACAL VOC 2012 test dataset. Our method \sens achieves comparable performance with the state-of-the-arts. Note that \cite{crfasrnn_iccv2015} used extra training data.}
\label{tab:seg_pascal_voc_2012}
\end{table}

\section{Conclusion}
Nonlinear structured learning has been a promising yet challenging topic in the community.
In this work, we have proposed a nonlinear structured learning method of tree potentials  for image segmentation. The unary and pairwise potentials are ensembles of class-wise trees, with the ensemble parameters and the trees jointly learned in a unified large-margin framework. In this way, nonlinearity is easily introduced into the \crf learning.
The resulted model involves exponential number of variables and constraints.
We therefore derive a novel algorithm combining a modified column generation method and the cutting-plane technique for efficient model training.
We have exemplified the superiority of the proposed nonlinear potential learning method by comparing against state-of-the-art methods on both binary and multi-class object segmentation datasets.
A potential disadvantage of our method is that it is prone to overfitting due to the outstanding non-linear learning capacity. This can be alleviated by using more training data.
On the other hand, as we show in Table \ref{tab:seg_msrc}, our method using pre-trained CNN features has shown the best performance.
Therefore it is worth exploiting to further combine our method with deep learning techniques in the future work.

\clearpage
{\small
	\bibliographystyle{ieee}
	\bibliography{s_ensmeble}
}

\end{document}